\documentclass[letterpaper, 10 pt, conference]{ieeeconf}  

\IEEEoverridecommandlockouts                              

\usepackage{graphicx}
\usepackage{amsmath,amsfonts}
\usepackage{algorithmic}
\usepackage{algorithm}
\usepackage{float}

\title{\LARGE \bf
Mobile Manipulation Planning for Tabletop Rearrangement
}

\author{Jiaming Hu \quad Jiawei Wang \quad Henrik I Christensen
}

\begin{document}

\maketitle
\thispagestyle{empty}
\pagestyle{empty}

\begin{abstract}

Efficient tabletop rearrangement planning seeks to find high-quality solutions
while minimizing total cost. However, the task is challenging due to object
dependencies and limited buffer space for temporary placements. The complexity
increases for mobile robots, which must navigate around the table with
restricted access. A*-based methods~\cite{orla} yield high-quality solutions, but
struggle to scale as the number of objects increases. Monte Carlo Tree Search
(MCTS) has been introduced as an anytime algorithm, but its convergence speed to
high-quality solutions remains slow. Previous work~\cite{strap2024} accelerated
convergence but required the robot to move to the closest position to the object
for each pick and place operation, leading to inefficiencies. To address these
limitations, we extend the planner by introducing a more efficient strategy for mobile robots. Instead of selecting the nearest available location for each
action, our approach allows multiple operations (e.g., pick-and-place) from a
single standing position, reducing unnecessary movement. Additionally, we incorporate
state re-exploration to further improve plan quality. Experimental results show
that our planner outperforms existing planners both in terms of solution quality
and planning time.

\end{abstract}

\section{INTRODUCTION}

Efficient tabletop rearrangement planning for mobile robots involves organizing objects into a desired arrangement while minimizing cost. Unlike stationary robots with full access to the workspace, mobile robots must navigate around the
table for pick-and-place actions when direct access is unavailable. Additionally, collisions often prevent objects from being moved directly to their goal positions, requiring action reordering based on object dependencies. Circular dependencies can lead to deadlocks, making temporary buffer allocation essential.

To address these challenges, \cite{trlb} proposed a scalable and fast planner
that expands its search tree across different states using lazy buffer
allocation. However, it often produces low-quality plans with redundant actions,
highlighting the need for more efficient planning. Monte Carlo Tree Search
(MCTS)-based anytime planners \cite{mcts2020labbe, mcts2024huang} improve plan
quality by refining solutions over time, but they converge slowly to
high-quality results. ORLA* \cite{orla} introduced tree search expansion with
lazy buffer allocation, achieving faster convergence, but it is not an anytime
planner and suffers from long planning times as the number of objects increases.

To overcome these limitations, our previous work, STRAP \cite{strap2024},
enhanced the A*-based approach by integrating goal-attempting and lazy buffer
allocation. This improvement enabled anytime planning with faster convergence,
balancing efficiency and solution quality.

\begin{figure}[t]
    \centering
    \includegraphics[width=0.4\textwidth]{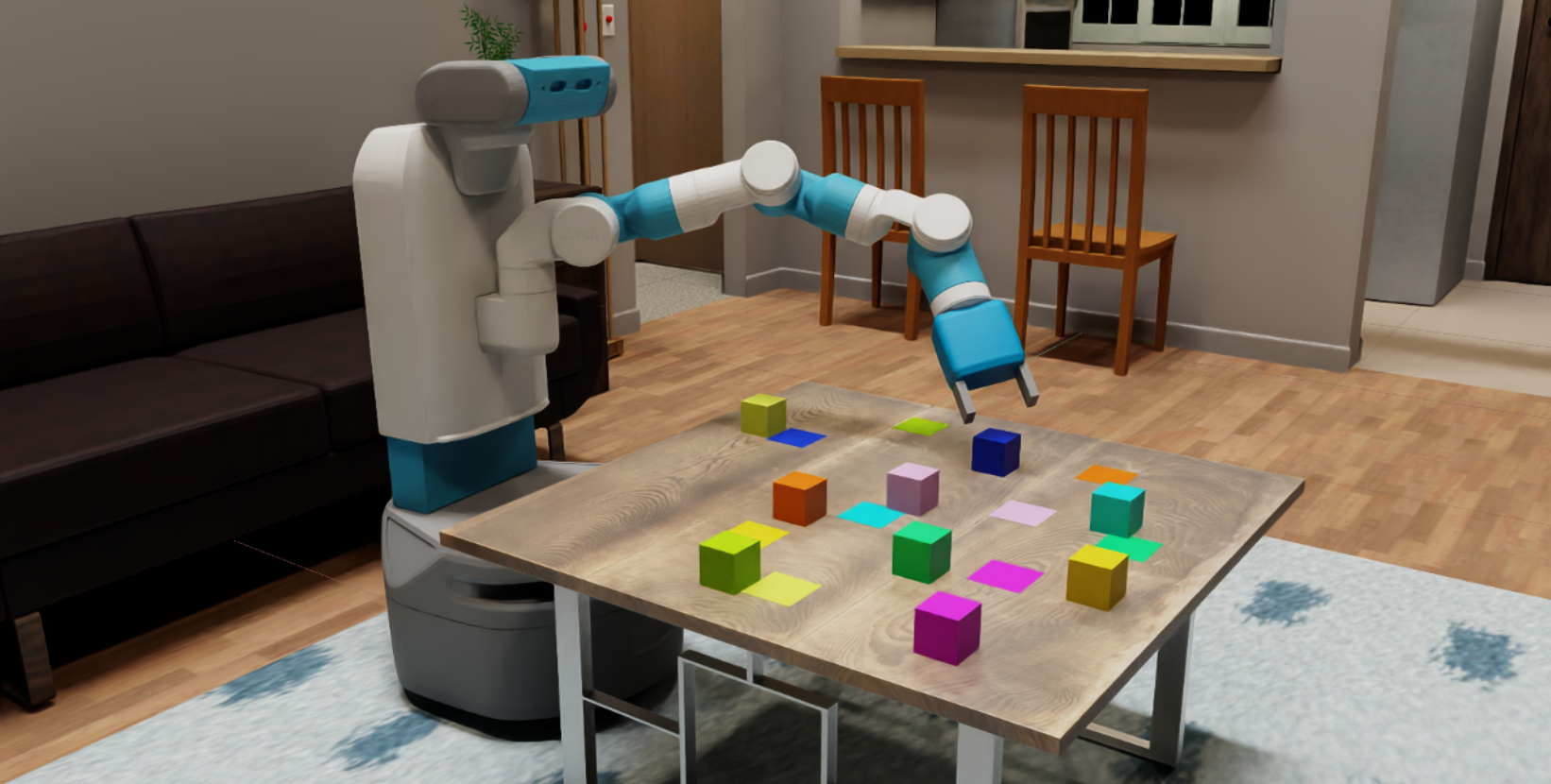}
    \caption{A mobile robot performs tabletop rearrangement, needing to move its
      base to access the entire table due to limited reach. Blocks of various
      colors are scattered and must be relocated to matching color positions. }
    \label{fig:simple_example}
\end{figure}

Both ORLA* and STRAP require the robot to move to the nearest location for each
pick-and-place action. While this simplifies decision-making, it results in
inefficient movement patterns. A more effective strategy would allow multiple
pick-and-place actions from a single standing position, reducing unnecessary movement.
However, this approach increases the number of possible actions from each state,
leading to greater complexity in search tree expansion.

We introduce an enhanced A*-based planner that improves efficiency by enabling
multiple pick-and-place actions from a single standing position. While this
increases the complexity of finding high-quality solutions, our approach
mitigates this challenge by employing an optimized expansion method that
selectively explores successor states, reducing unnecessary searches.
Additionally, we incorporate state re-exploration to refine plan quality over
time. The following sections review related work, describe our proposed
algorithm, and present experimental results demonstrating its advantages over
existing planners.

\section{Related Work}
Rearrangement planning are categorized as prehensile~\cite{trlb, mcts2020labbe,
  mcts2024huang, orla, runningbuffer}, non-prehensile~\cite{dogar2011framework}, and hybrid~\cite{pmlr-v205-tang23a}. This paper focuses on
prehensile methods using pick-and-place.

\subsection{Rearrangement Planning with External Buffer}
Early works~\cite{krontiris2015rss, krontiris2016icra} utilized a dependency graph
to model object relationships. However, potential deadlocks may
cause the problem to be infeasible, necessitating the use of temporary placements 
to resolve such issue ~\cite{han2018ijrr}.
Nevertheless, external buffers are a limitation for general cases, a solution is 
many simultaneous storage spaces, or ``running buffers''~\cite{runningbuffer}, which 
is ineffective for mobile manipulation. 

\subsection{Rearrangement Planning with Internal Buffer}
External buffers may be unavailable in practice, prompting the development of
TRLB~\cite{trlb}, which focuses on use of internal buffers. With lazy buffer
allocation, it effectively handles more objects, but produces
lower-quality solutions. Better solutions can be generated using MCTS-based approaches~\cite{mcts2020labbe, mcts2024huang}. The MCTS planner is 
anytime, but it has a slow convergence speed. A refinement was
ORLA*~\cite{orla}, which utilizes lazy buffer allocation, and prioritized states. 
While ORLA* demonstrated better plan quality, its state representation
only considers the object arrangements excluding robot locations, which leads to
sub-optimality. Our previous work~\cite{strap2024} considers the robot location
and provides the anytime property by 
incorporating goal-attempting strategies.

\subsection{Learning-Based Approach}
Reinforcement learning (RL) has been employed to address the complexity of
rearrangement planning.~\cite{qureshi2021rss} proposed a neural rearrangement
planner using deep RL to recursively solve rearrangement tasks.
~\cite{ghosh2022ijcnn} developed a deep RL-based task planning method for
large-scale object rearrangement, demonstrating generalizability across
different scenarios.~\cite{chen2024} proposed a combined task-level RL and
motion planning framework, leveraging distributed deep Q-learning and an
A*-based post-processing technique to improve efficiency. Nevertheless, the
success rates were limited as the number of objects increased.

\begin{figure}[t]
    \centering
    \includegraphics[width=0.4\textwidth]{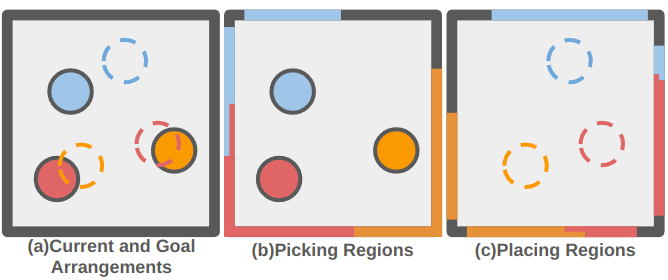}
    \caption{(a) The current state's arrangement is represented by solid circles and goal arrangement indicated by a dashed circle. (b) Picking regions. For each object, the highlighted table side in the same color indicates where the robot can stand to pick it up. When regions overlap, the colors are evenly divided, with each occupying half of the space. (c) Placing regions. For each object, the region with the same color on the table side indicates the region where the robot can stand to place. }
    \label{fig:regions}
\end{figure} 

\section{Problem Statement}\label{problem_statement}

Given $n$ objects, an arrangement is defined as $\{p_1,..., p_n\}$, where each $p_i$ is the object pose, and it is valid if no object is in collision or out of boundary. This study does not consider stacking objects. A rearrangement plan is a sequence of relocation actions $[a_1, a_2, ...]$, where each action $a_i$ is an operation sequence. Each operation is defined as $(t, k, p, l)$ where $t$ is the operation type such as pick or place, $k$ denotes the manipulated object, $p$ represents the object pose for manipulation, $l$ specifies the robot's standing location. When the robot executes an operation, it must reset and move its base to the operation’s standing location if it differs from the current standing location. Otherwise, the robot can execute the operation directly without resetting or moving base. For example, given an operation sequence [$(pick, k_1, p_{pick_1}, l_{pick_1})$, $(place, k_1, p_{place_1}, l_{place_1})$], if $l_{pick_1}=l_{place_1}$, the robot can position its base at $l_{pick_1}$ and move the object $k_1$ directly from $p_{pick_1}$ to $p_{place_1}$. Otherwise, after picking $k_1$, the robot must reset the arm and move its base to $l_{place_1}$ for placing $k_1$ at $p_{place_1}$. Similarly, if an additional operation is added, such as [$(pick, k_1, p_{pick_1}, l_{pick_1})$, $(place, k_1, p_{place_1}, l_{place_1})$, $(pick, k_2, p_{pick_2}, l_{pick_2})$] where $l_{place_1}=l_{pick_2}$, then the robot can directly pick the object $k_2$ after placing $k_1$ in standing location $l_{place_1}$. Furthermore, for each operation $op$, the robot must be able to reach its manipulation location $p$ from its standing position $l$.
The goal is searching for an action sequence to rearrange from a start arrangement to a goal arrangement with the minimum total cost. Notably, each object should initially not in its goal pose.

Two distinct manipulation strategies define actions differently. The first one is ``single relocation action strategy'' used in \cite{orla,strap2024}. Given a single relocation action $a_i$ to move object $k_{a_i}$ from $p^{a_i}_{pick}$ to $p^{a_i}_{place}$, $l^{a_i}_{pick}$ and $l^{a_i}_{place}$ are the closest locations around the table to $p^{a_i}_{pick}$ and $p^{a_i}_{place}$ respectively. After moving base to $l^{a_i}_{pick}$, the robot picks the $k_{a_i}$, then resets the arm and moves base to $l^{a_i}_{place}$ for placing $k_{a_i}$ at $p^{a_i}_{place}$. Therefore, $a_i$ is defined as [($pick, k_{a_i}, p^{a_i}_{pick}, l^{a_i}_{pick}$), ($place, k_{a_i}, p^{a_i}_{place}, l^{a_i}_{place}$)]. Given the action sequence $[a_1, ..., a_m]$, its total cost will be
\[2\cdot MC \cdot m + TC(l_{robot}, l^{a_1}_{pick}) + \]
\[TC(l^{a_1}_{pick}, l^{a_1}_{place}) + \sum_{i=2}^{m}TC(l^{a_{i-1}}_{place}, l^{a_i}_{pick}) + TC(l^{a_i}_{pick}, l^{a_i}_{place})\]
where $MC$ is the manipulation cost while $TC$ is the travel cost. $l_{robot}$ is the initial robot location around the table. Akin to \cite{orla}, $TC(l_{1}, l_{2})$ is the total Euclidean distance for the robot from $l_{1}$ to $l_{2}$ around the table. On the other hand, $MC$ is the manipulation cost associated with pick and place operations.

Alternatively, the second strategy, termed the ``multiple relocation action strategy'', enables the robot to perform multiple picking or placing in a single standing location. Each multiple relocation action $a_i$ consists of a sequence of standing locations $[l^{a_i}_1, l^{a_i}_2, ...]$, while each standing location $l^{a_i}_j$ is associated with a set of operations. Between two adjacent operations with different standing locations, the robot must reset the arm with a grasped object and move its base to the next standing location. If the robot can reset without a grasped object, then these two operations should not be part of the same action. Therefore, the total cost of an action sequence $[a_1, ..., a_m]$ is
\[TC(l_{robot}, l^{a_1}_1) + C(n^{l^{a_1}_1}) + \sum_{j=2}^{v^{a_1}}[ TC(l^{a_1}_{j-1}, l^{a_1}_j) + C( n^{l^{a_1}_j})] +\]
\[\sum_{i=2}^{m}\{TC(l^{a_{i-1}}_{v^{a_{i-1}}}, l^{a_{i}}_1) + C(n^{l^{a_i}_1}) +\sum_{j=2}^{v^{a_i}}[ TC(l^{a_i}_{j-1}, l^{a_i}_j) + C(n^{l^{a_i}_j})]\}\]
where $n^{l_j^{a_i}}$ is the number of operations performed at $l_j^{a_i}$ and $C(n) = MC \cdot n$, while $v^{a_i}$ is the number of standing locations involved in $a_i$.

\begin{figure}[!t]
\centering
\includegraphics[width=3.2in]{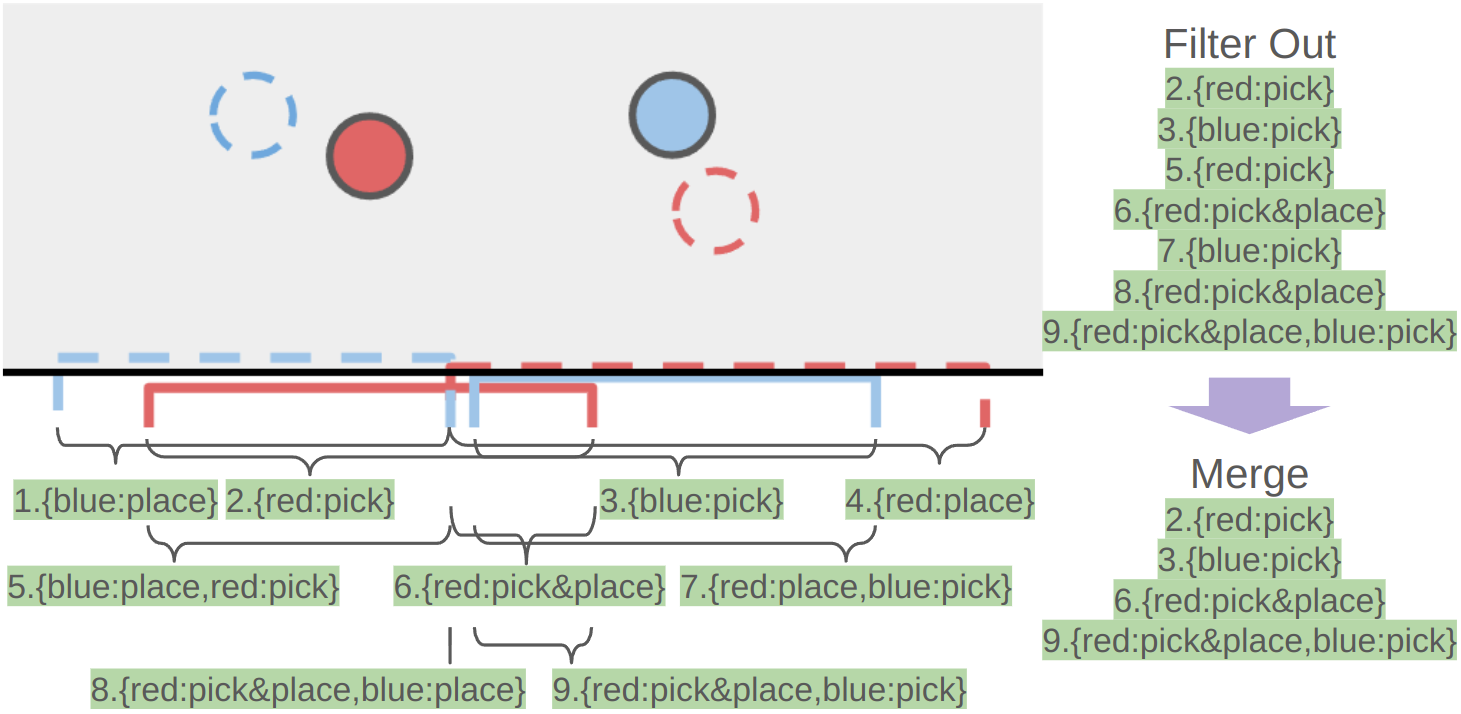}
\caption{The process of generating manipulation regions, along with their corresponding object-operation dictionary defining the feasible operations on objects within each region. Left: Based on placing and picking regions, find initial manipulation regions with their object-operation dictionary on one table side, represented as a set of indexed green boxes. Right: The filtering process removes non-manipulable objects from the object-operation dictionary and eliminates regions if they become empty. The merging process then combines the filtered regions.}
\label{fig:manipulation_regions}
\end{figure}

\section{Search Tree with Different Strategies}
Using a search tree to explore different states in pursuit of a high-quality plan requires a definition of how each state is expanded into its successor states.

\subsection{State Representation}
A state is characterized by object poses and the robot’s position on the table side, expressed as $(l_{robot}, p_1, p_2, ...)$, where $l_{robot}$ denotes the robot's standing location around the table, and each $p_i$ corresponds to the pose of an individual object $i$. Notably, the robot's arm is positioned in its default configuration without grasping anything in each state.

\subsection{Tree Expanding Actions}
The expanding process searches for successor states from a given state, determined by the manipulation strategy.

\subsubsection{Single Relocation Action Strategy}

A single relocation action consists of moving to the closest position around the table to the selected object, picking it up, and then moving to the closest position to the selected placing location around the table for placing. Given a state, if the object's goal position is unoccupied, placing it at its goal results in a successor state. If the goal position is occupied, multiple successor states arise based on allocated feasible buffer positions by sampling. No successor state is generated if the object is already at its goal.

\subsubsection{Multiple Relocation Action Strategy} \label{sec:multiple_relocation}
A multiple relocation action strategy allows multiple pick-and-place on a single standing position. Given a state, we identify all associated picking and placing regions to define the manipulation regions. Each manipulation region is associated with a set of standing locations, while each standing location can perform different operation sequences, which in turn, leads to a set of successor states originating from the given state. The following section elaborates on these processes in detail.

Given the current state's arrangement $A_{current}$, both the picking and placing regions are determined based on both $A_{current}$ and $A_{goal}$. The picking and placing regions are a set of table side regions for picking and placing objects respectively. 
Picking regions are defined by $A_{current}$, as shown in (b) of Fig.\ref{fig:regions}, while placing regions are determined by $A_{goal}$, as illustrated in (c) of Fig.\ref{fig:regions}.

The next step is estimating the manipulation regions of the current state, and it is formed from its picking and placing regions, as well as their intersections on the table, as shown on the left of Fig.~\ref{fig:manipulation_regions}. Each manipulation region is associated with a object-operation dictionary containing a set of objects with possible operations; however, not all objects within each region are manipulable. An object is manipulable in a region if it can be picked up; if it can only be placed, it is not manipulable. A filtering process removes non-manipulable objects from the region. If, after filtering, no objects remain in a manipulation region, that region will be deleted. The merging process combines two regions if they share the same object-operation dictionary. Consequently, we obtain a set of manipulation regions, each associated with an object-operation dictionary, as illustrated in the bottom right of Fig.~\ref{fig:manipulation_regions}. 

However, with increasing number of objects, the number of manipulation regions related to a state will increase exponential. To address this, we introduce a region reduction strategy: for any location of a manipulation region $R$, if another region provides equal or greater operational coverage for that location, then $R$ is ignored. For example, consider three regions with their object-operation dictionaries: $R_1$ with $\{blue:pick\}$, $R_2$ with $\{blue:pick,red:pick\}$, and $R_3$ $\{blue:pick,yellow:pick\}$. Any location of $R_1$ is in either $R_2$ or $R_3$. Even though neither $R_2$ nor $R_3$ fully encompasses $R_1$, $R_1$ is disregarded in this case. This strategy reduces the number of regions requiring consideration. Later, in Sec.~\ref{state_reduction}, we further analyze region reduction strategy.

\begin{figure}[!t]
\centering
\includegraphics[width=3.0in]{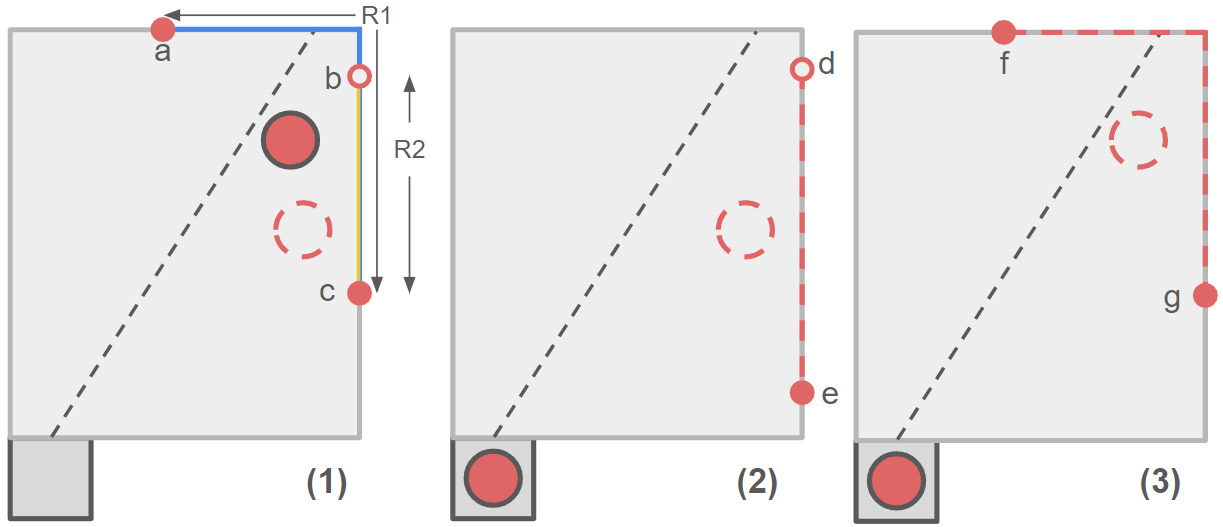}
\caption{The left-bottom box is the robot. (1) In a given state, there are two manipulation regions, R1 and R2. For R1, points $a$ and $c$ are efficient standing locations. For R2, only point $c$ is efficient because it lies on the shortest path from the robot to point $b$. (2) When the object is in hand and the placing region is on the right side, only point e is the feasible placing point. (3) When the object is in hand and the placing region is on both the top and right side, both f and g are feasible placing points.}
\label{fig:placing_points}
\end{figure}

Efficient standing locations are identified based on the manipulation regions of the current state. Different regions enable specific operations on objects, and once a region is selected, its optimal standing location is determined. To minimize travel distance, only the endpoints of manipulation regions are considered as efficient standing locations. Thus, for each manipulation region, efficient standing locations are limited to its endpoints. However, as shown in (1) of Fig~\ref{fig:placing_points}, not all endpoints of the manipulation regions are considered as efficient standing locations. If the shortest path from the robot position of the current state to one endpoint $p_1$ passes through another endpoint $p_2$ within the same manipulation region, the former is ignored. This is because any operation sequence involving $p_1$ can be substituted with one using 
$p_2$ for a lower cost. 

Given a standing location, searching for its feasible operation sequences is essential. We first distinguish its explicit and implicit object set. Explicit objects require the robot to pick them up and move the base for later placement, whereas implicit objects can be picked and placed while standing in that location. Thus, upon moving to a standing point, the robot first rearranges all implicit objects and then selects one (or none) explicit object to grasp before moving base. According to the manipulation region $R$ of the standing location $l$, we identify a set of objects along with their associated operations. Without considering collision, if an object can only be picked up in $R$ but not placed, it is classified as an explicit object for $l$. Conversely, if the robot can pick and place the object within the region $R$, this object is considered a potential implicit object for $l$. 
Then, it is essential to verify whether an object is truly implicit for standing location $l$. Given the current and target arrangements of all potential implicit objects $l$, we first calculate an abstract plan minimizing the running buffer, as described in~\cite{runningbuffer1}, without initially allocating buffer space. That is, this abstract plan is a sequence of abstract action moving an object to its goal or buffer, such as $[(k_1, \text{`to goal'}), (k_2, \text{`to buffer'}), ...]$. Then, the buffer allocation function of~\cite{trlb} can verify the feasibility of the abstract plan by sampling feasible buffer placements for those ``to buffer'' actions. However, explicit objects may occupy the goal positions of certain potential implicit objects. To address this, if an abstract action $a$ is moving an object to its goal location occupied by an explicit object, we insert an abstract action to move that explicit object to a buffer before $a$ in the abstract plan. Subsequently, the buffer allocation function~\cite{trlb} solves the updated abstract plan. If successful, all potential implicit objects are confirmed as implicit for $l$, and a corresponding rearrangement plan is produced. If no solution is found, the potential implicit object that caused the buffer allocation failure is reclassified as explicit. The minimal buffer plan for the remaining potential implicit objects is recalculated, and buffer allocation is attempted again. This iterative process continues until a solution is found or no potential implicit objects remain. Ultimately, the system identifies both explicit and implicit objects of each standing location with a rearrangement plan for those implicit objects. Importantly, standing locations within the same manipulation region share a potential implicit object set, though their actual implicit object sets may differ due to variations in available buffer.

\begin{figure}[!t]
\centering
\includegraphics[width=3.4in]{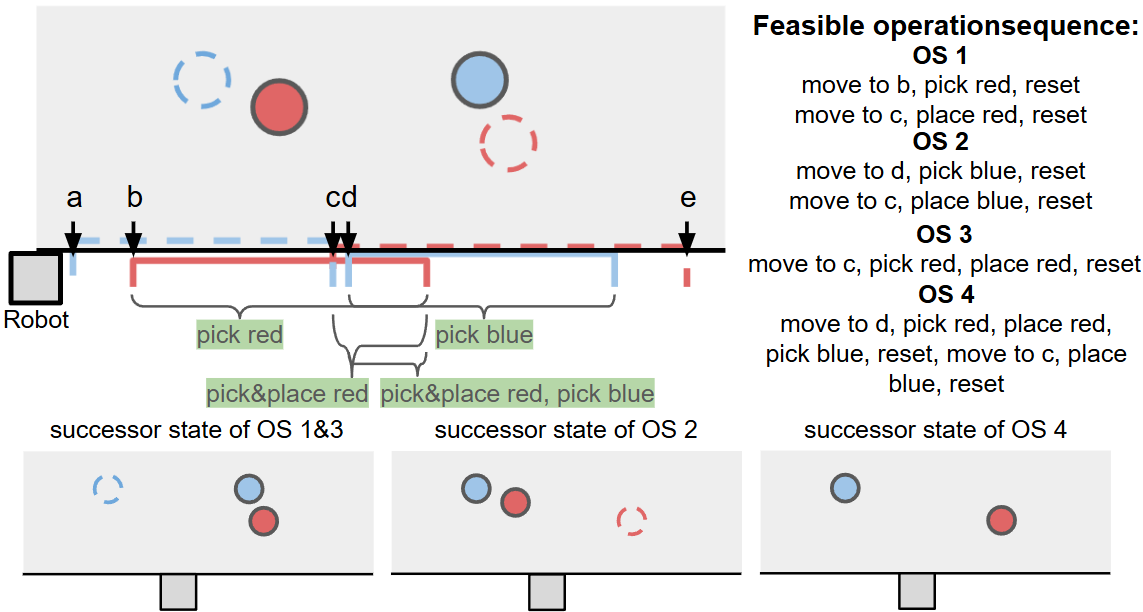}
\caption{Given a state where the robot only moves on one table side, there are 4 successor states, shown in the bottom, lead by 4 feasible operation sequences from the current state.}
\label{fig:state_action}
\end{figure}

If the standing location contains no explicit objects, its only operation sequence involves rearranging all implicit objects resulting in a single successor state. Conversely, if explicit objects are present,  a decision must be made on whether to grasp one and, if so, which object to grasp for subsequent placement. After selecting an explicit object, its placing location must be determined. If the goal location of the selected object is not occupied, then this goal location will be its placing location. If not, the placing location is randomly sampled from collision-free space. Given the placing location, its placing region around the table is calculated. If the current standing location falls within this region, the explicit object can be placed directly without moving the base. Otherwise, the endpoints of the placing region become the potential standing locations for placing the object. However, not all endpoints are utilized; a standing point for placing is considered if the shortest path to it does not pass through other endpoints of the same placing region, as shown in (2) and (3) of Fig.~\ref{fig:placing_points}. Eventually, an operation sequence is determined, leading to a new successor state.

\begin{figure}[!t]
\centering
\includegraphics[width=3.5in]{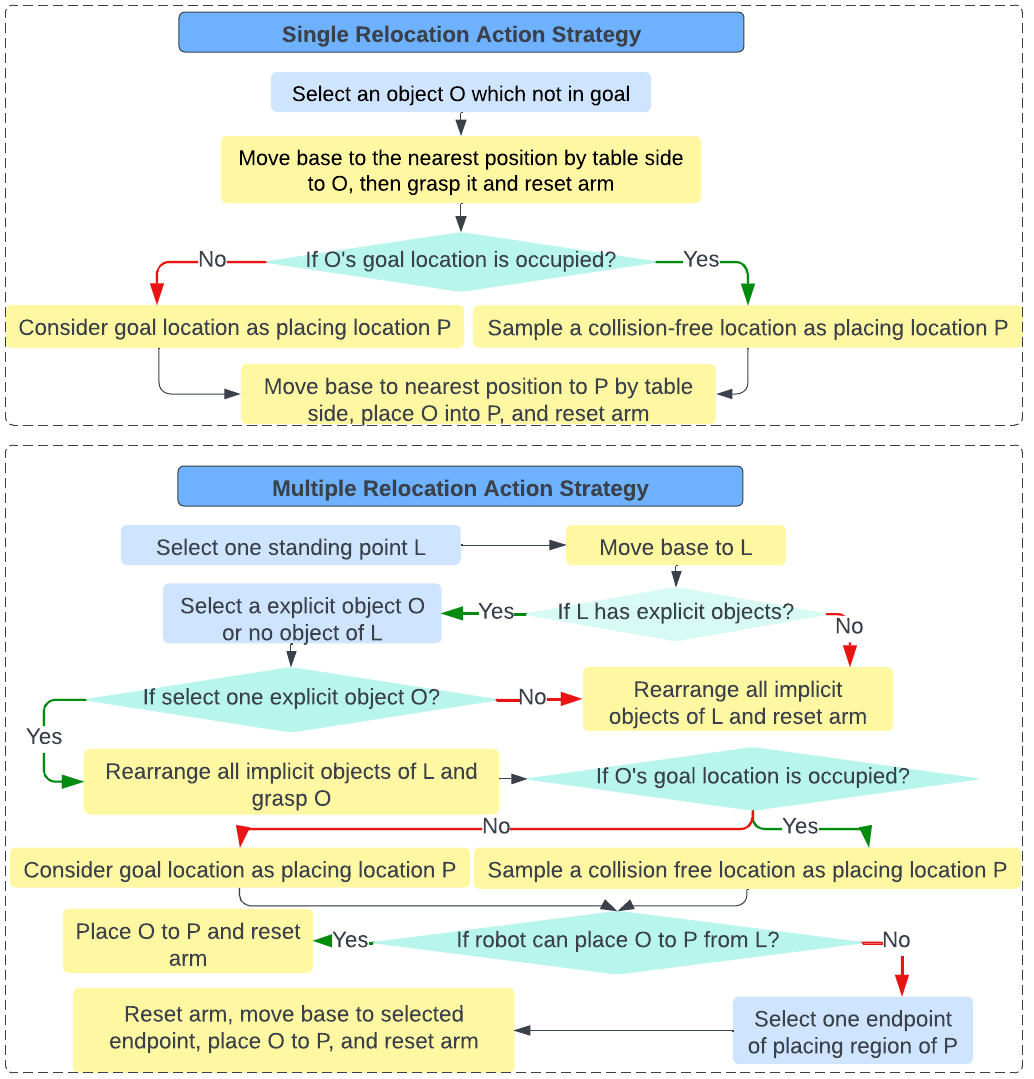}
\caption{The flowcharts of two strategies to generate an operation sequence from a state to its successor. By following each flowchart, an operation sequence is produced. Selection blocks (light blue boxes) represent decision points, condition blocks (light green diamonds) determine the next step based on specific conditions, and process blocks (yellow boxes) indicate operations the robot must perform. }
\label{fig:flowchart}
\end{figure}

For clarity, the flowchart in Fig.~\ref{fig:flowchart} illustrates the generation of an operation sequence. Various decision combinations on the flowchart yield the operation sequence from a given state to its successor state based on action strategy.

Based on Fig.~\ref{fig:flowchart}, multiple relocation action has pattern: 
\[MB + DO + [MBWO + PO]\cdot y + RESET\]where $MB$ is moving base without object, $DO$ is doing a set of operations such as picking or placing, $MBWO$ is moving base with an object, $PO$ is placing object, and $RESET$ is resetting arm. Furthermore, $MBWO + PO$ is optional here, so $y$ can be zero or one. However,  this pattern is more accurately represented as: \[MB + [DO + MBWO]\cdot x + DO + [MBWO + PO]\cdot y + RESET\]
where $x$, referred to as the ``manipulation loop count,'' can be any natural number. Since $x$ can be large or even infinite, it is impractical to generate operation sequences that enumerate all possible values of $x$ during expanding from a given state.

To address this problem, we propose a shortening function. Given a state $S_{current}$ along with its parent state $S_{parent}$ and child state $S_{child}$, the function evaluates whether a shorter operation sequence, $OP_{S_{parent}\text{ to } S_{child}}$, can be formed by combining the operation sequences $OP_{S_{parent}\text{ to } S_{current}}$ and $OP_{S_{current}\text{ to } S_{child}}$. If such a sequence is found, it enables a direct connection from $S_{parent}$ to $S_{child}$ using $OP_{S_{parent}\text{ to } S_{child}}$. This shortening process involves two steps: switch-and-cancel and operation merging.

The switch-and-cancel concept involves reordering the operations in both $OP_{S_{parent}\text{ to } S_{current}}$ and $OP_{S_{current}\text{ to } S_{child}}$ so that the last placing operation of $OP_{S_{parent}\text{ to } S_{current}}$ and the first picking operation of $OP_{S_{current}\text{ to } S_{child}}$ apply to the same object, making these two operations redundant. Then, we shorten the operation sequences by removing these redundant operations and then concatenating the remaining sequences. Given $OP_{S_{parent}\text{ to } S_{current}}$ in the format
\[[..., op_1, op_2] = [..., (pick, k_1, p_1, l_1), (place, k_1, p_2, l_2)]\]
and $OP_{S_{current}\text{ to } S_{child}}$ in the format
\[[op_3, op_4, ...] = [(pick, k_2, p_3, l_3), (place, k_2, p_4, l_4), ...]\]
we define two subsequences. The first, $OP^{post\text{ }pick-n-place}_{S_{parent}\text{ to } S_{current}}$, is a list of pick-and-place operations that share the same standing location as the last operation in $OP_{S_{parent}\text{ to } S_{current}}$, and it is cropped from the end of $OP_{S_{parent}\text{ to } S_{current}}$. The second, $OP^{pre\text{ }pick-n-place}_{S_{current}\text{ to } S_{child}}$, is a list of pick-and-place operations that share the same standing location at the first operation in $OP_{S_{current}\text{ to } S_{child}}$ and is cropped from the beginning of $OP_{S_{current}\text{ to } S_{child}}$.
Then, we can shorten sequences by the following conditions:
\begin{itemize}
  \item $l_1 \neq l_2 \And l_3 \neq l_4$: If $k_1 = k_2$, remove $op_2$ and $op_3$, then concatenate two operation sequences.
  
  \item $l_1 \neq l_2 \And l_3 = l_4$: We locate the first pick-and-place operations, $op_5$ and $op_6$, that apply to $k_1$ from $OP^{pre\text{ }pick-n-place}_{S_{current}\text{ to } S_{child}}$. We then attempt to switch them to the beginning of $OP_{S_{current}\text{ to } S_{child}}$ without causing the operation sequence infeasible. If this is possible, then we can remove $op_2$ and $op_5$, then concatenate two sequences.
  
  \item $l_1 = l_2 \And l_3 \neq l_4$: We locate the last pick-and-place operations, $op_5$ and $op_6$, that apply to $k_2$ from $OP^{post\text{ }pick-n-place}_{S_{parent}\text{ to } S_{current}}$. Then attempt to switch them to the end of $OP_{S_{parent}\text{ to } S_{current}}$ without causing the operation sequence infeasible. If this is possible, then remove $op_6$ and $op_3$ and concatenate two sequences.
  
  \item $l_1 = l_2 \And l_3 = l_4$: For each common object $k_{common}$ used in $OP^{post\text{ }pick-n-place}_{S_{parent}\text{ to } S_{current}}$ and $OP^{pre\text{ }pick-n-place}_{S_{current}\text{ to } S_{child}}$, we locate the last pick-and-place operations, $op_5$ and $op_6$, on $k_{common}$ from $OP_{S_{parent}\text{ to } S_{current}}$ and first pick-and-place operations, $op_7$ and $op_8$, on $k_{common}$ from $OP_{S_{current}\text{ to } S_{child}}$. We attempt to switch $op_5$ and $op_6$ to the end of $OP_{S_{parent}\text{ to } S_{current}}$ and $op_7$ and $op_8$ to the start of $OP_{S_{current}\text{ to } S_{child}}$ without making either sequence infeasible. If this is possible, then remove $op_6$ and $op_7$ and concatenate two sequences.
\end{itemize}

If switch-and-cancel doesn't shorten two sequences, the operation merging is applied. The idea of merging is combining operations with one standing location instead of two standing locations. First, we find two sub sequences, $OP^{post\text{ }operations}_{S_{parent}\text{ to } S_{current}}$ and $OP^{pre\text{ }operations}_{S_{current}\text{ to } S_{child}}$, from $OP_{S_{parent}\text{ to } S_{current}}$ and $OP_{S_{current}\text{ to } S_{child}}$ respectively. $OP^{post\text{ }operations}_{S_{parent}\text{ to } S_{current}}$ is a list of operation cropped from the back of $OP_{S_{parent}\text{ to } S_{current}}$, where all operations share the same standing location. Similarly, $OP^{pre\text{ }operations}_{S_{current}\text{ to } S_{child}}$ is a list of operation cropped from the beginning of $OP_{S_{current}\text{ to } S_{child}}$, also sharing the same standing location. Importantly, $OP^{post\text{ }operations}_{S_{parent}\text{ to } S_{current}}$ and $OP^{pre\text{ }operations}_{S_{current}\text{ to } S_{child}}$ are distinct from $OP^{post\text{ }operations}_{S_{parent}\text{ to } S_{current}}$ and $OP^{pre\text{ }operations}_{S_{current}\text{ to } S_{child}}$. For instance, consider the operation sequence
[$(pick, k_1, p_1, l_1)$, $(place, k_1, p_2, l_2)$, $(pick, k_2, p_3, l_2)$, $(place, k_2, p_4, l_2)$], its $OP^{post\text{ }operations}$ is the last three operations, while its $OP^{post\text{ }pick-n-place}$ is only the last two operations. Given the standing location $l$ of $OP^{post\text{ }operations}_{S_{parent}\text{ to } S_{current}}$, we check if the robot can complete all operations of $OP^{pre\text{ }operations}_{S_{current}\text{ to } S_{child}}$ from $l$. If so, we will update the standing location of $OP^{pre\text{ }operations}_{S_{current}\text{ to } S_{child}}$ to $l$ and concatenate two sequences. Similarly, given the standing location $l$ of $OP^{pre\text{ }operations}_{S_{current}\text{ to } S_{child}}$, we will check if the robot can complete all operations of $OP^{post\text{ }operations}_{S_{parent}\text{ to } S_{current}}$ from $l$. If so, we will update the standing location of $OP^{post\text{ }operations}_{S_{parent}\text{ to } S_{current}}$ to $l$ and concatenate two sequences.

As a result of these shortening processes, states can have an operation sequence with a manipulation loop count greater than zero that leads to its successor state. Over time, this approach enables each state to naturally explore operation sequences with varying manipulation loop counts.

\section{A*-based Planning for High-quality plan}
This section discusses how the search tree explores states to achieve a high-quality plan. The previous work, STRAP, applies an A* based approach using a heuristic function to prioritize which states to explore. However, due to the continuous nature of the workspace, there exists an infinite buffer location option for object placement, resulting in each state having an unbounded set of possible successor states. To address this, STRAP limits the number of actions involving moving objects to the buffer from any given state. This parameter significantly impacts overall performance. To manage this, re-exploring states becomes essential, necessitating a rewrite function to maintain tree consistency. Additionally, our algorithm must account for multiple relocation action strategies, requiring modifications to the goal-attempting.

\begin{figure}[!t]
\centering
\includegraphics[width=3.5in]{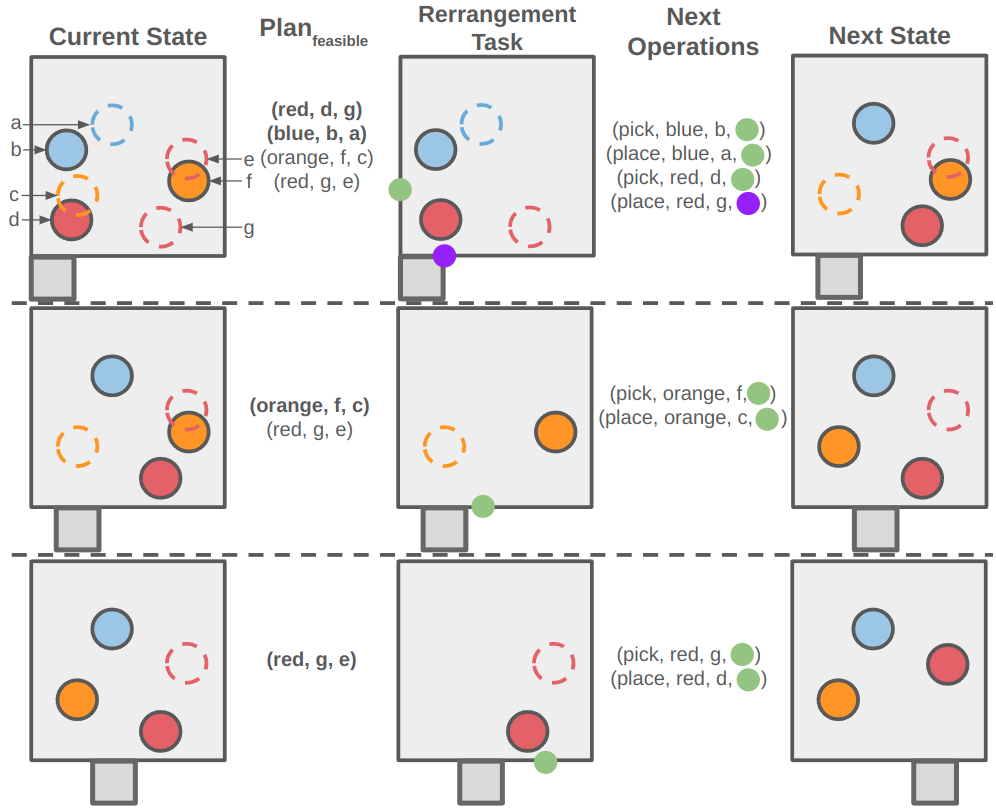}
\caption{An example of goal attempting. In each iteration, we identify the initial portion (bold) of $Plan_{feasible}$ that can be executed directly. Based on that initial portion, we define a rearrangement task and seek a standing point (green dot on the table side) that maximizes the number of operations. If the last operation is a pick action, the robot must move to the nearest standing point (purple dot on the table side) to place the object. Then, we remove the completed actions from $Plan_{feasible}$.}
\label{fig:attempt_to_goal}
\end{figure}

\subsection{Main Pipeline}
The pipeline follows the A* principle, exploring states based on a $g$ value for the cost from the start and an $h$ value as a lower bound to the goal, with the state minimizing $g + h$ selected for exploration; our work uses the same heuristic function from \cite{orla} depending on arrangement difference. Our notable distinction is allowing re-explore states (in the closed list) with a certain probability. Once a state is selected for exploration, the exploration step will identify successor states based on the manipulation strategy. If any successor state has been previously explored (exists in the closed list) with a lower $g$ value, a tree-rewriting function is applied. Following this, a goal-attempting function is used to generate a feasible path from the current state to the goal. If the total cost of the combined plan concatenates the path from the start to the current state with this feasible solution to the goal is lower than the current best plan, the best plan is updated accordingly. This iterative process continues until a timeout.

\subsection{Exploration Step}
Exploration step is searching for the successor states of the given state and adding them to the open list. For each state, we enumerate possible selections in the flowchart (illustrated by light blue blocks in Fig.~\ref{fig:flowchart}) to generate operation sequences, with each sequence leading to a successor state. Under a single relocation action strategy, the only selection involves which object that is not in its goal position to manipulate. In contrast, a multiple relocation action strategy involves several selections in the flowchart, requiring enumeration of all combinations to generate different successor states.
Furthermore, if the current explored state exists in the closed list with a lower $g$ value, tree rewriting is required, then the exploration step is skipped for that state.

\subsection{Tree Rewriting}
Tree rewriting is necessary when state re-exploration reveals a lower-cost path to the same state. As shown in Alg.~\ref{alg:rewrite}, the $Rewrite$ function will remove the connection between the explored state $S_{explored}$ with its current parent and connect to the new parent $S_{parent}$. After that, it will update the $g$ value and best action sequence to each state of the sub-tree under $S_{explored}$. Then, for each child state $S_{child}$ of $S_{explored}$, we need to check if it can be achieved directly from $S_{parent}$.  If so, the child is reconnected directly to $S_{parent}$, followed by updating its sub-tree. For single relocation strategy, if the last action from $S_{parent}$ to $S_{explored}$ applies on the same object of the operation sequence from $S_{explored}$ to $S_{child}$, then we can reconnect from $S_{parent}$ to $S_{child}$ by operation sequence of moving that object from its location in $S_{parent}$ to its location in $S_{child}$. For multiple relocation action strategy, the process is similar to the shortening method in Section.~\ref{sec:multiple_relocation}. 

\begin{algorithm}[H]
\caption{Rewrite}\label{alg:rewrite}
\begin{algorithmic}[1]
\STATE \textbf{Input:} Parent State $S_{parent}$, Explored State $S_{explored}$
\STATE $searchT.RemoveEdge(S_{explored}, S_{explored}.parent)$
\STATE $searchT.Connect(S_{parent}, S_{explored})$
\STATE $UpdateSubtree(S_{explored})$
\FOR{$childS \in S_{explored}.children$}
\IF{$S_{parent}.IsAchievableTo(childS)$}
\STATE $searchT.RemoveEdge(S_{explored}, childS)$
\STATE $searchT.Connect(S_{parent}, childS)$
\STATE $UpdateSubtree(childS)$
\ENDIF
\ENDFOR
\end{algorithmic}
\end{algorithm}

\subsection{Goal Attempting}
Goal attempting involves quickly searching for a feasible rearrangement plan from the current state to the goal, providing the anytime property. As in~\cite{strap2024}, goal attempting for single relocation actions is relatively straightforward. However, coordinating multiple relocation actions is more complex. For the multiple relocation strategy, the goal attempting must still ensure both fast planning and a high success rate.
To achieve this, we first calculate a feasible rearrangement sequence, $Plan_{feasible}$, using the rearrangement local solver with lazy buffer allocation from ~\cite{trlb}, ensuring a high success rate. The format of $Plan_{feasible}$ is as follows:
\[[(k_1, p^{k_1}_{pick}, p^{k_1}_{place}),(k_2, p^{k_2}_{pick}, p^{k_2}_{place}), ...]\]where $k_i$ is the object while $p^{k_i}_{pick}$, $p^{k_i}_{place}$ are the pick and place locations of $k_i$.
At this moment, $Plan_{feasible}$ does not account for the robot’s standing locations. We identify the initial portion of $Plan_{feasible}$ which can be performed directly without collision. The initial portion defines a rearrangement task. After calculating the manipulation regions of the initial state of this rearrangement task, we identify the closest standing point where the robot can perform the maximum number of operations. If needed, the robot may move to place the final object. Once these operations are completed, the corresponding completed actions are removed from $Plan_{feasible}$. This cycle repeats until the $Plan_{feasible}$ is exhausted. This approach is greedy, so a solution can be produced quickly. An example is demonstrated in Fig.~\ref{fig:attempt_to_goal}.

\section{Evaluation}
\begin{figure}[!t]
\centering
\includegraphics[width=2.6in]{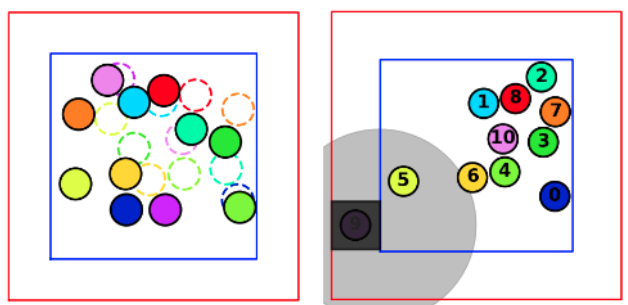}
\caption{The large red box represents the room, and the blue box indicates the table. Left: The initial rearrangement task setup. Right: The robot is moving an object to a new standing location for placement. The gray circle denotes the manipulation range of the robot.}
\label{fig:experiment}
\end{figure}

This evaluation primarily compare the performance of various planners and manipulation strategies on different rearrangement tasks for a mobile robot. To ensure fair comparisons, we utilize the disk experiment setup from~\cite{strap2024, orla, trlb}, as shown in Fig.~\ref{fig:experiment}. we define the robot's manipulation range as half of the smaller dimension between the table's width and height, as $0.5\cdot \min\{width, height\}$, allowing the robot to access the entire table surface by moving around it. Notably, the experiment setup does not permit stacking objects on top of one another.

The experiments compare the following planners:
\begin{itemize}
  \item TRLB: Planner~\cite{trlb} using bidirectional search tree with lazy buffer allocation.
  \item ORLA*: A* planner~\cite{orla} using lazy buffer allocation.
  \item STRAP: A* planner~\cite{strap2024} with goal attempting using lazy buffer allocation.
  \item STRAP V2: Improved STRAP with re-exploration.
  \item MCTS: MCTS planner using reward function from~\cite{mcts2024huang}.
\end{itemize}
Since most of the planners possess the anytime property, we will demonstrate their cost over the time. The cost is defined as in~\ref{problem_statement}, with the manipulation cost initially set to 1. For each setup, we run the planners with various start and goal arrangements, recording the solution cost at each time interval. After 100 trials, we compute the average path cost for each time period during execution. Additionally, we set a sufficiently high timeout for this evaluation to ensure a 100\% success rate across all algorithms. In STRAP V2, state re-exploration occurs after the goal is achieved. Once achieved, there is a 30\% probability of re-exploring a state in each subsequent iteration.

\subsection{Evaluation on Different Relocation Action Strategy}

\begin{figure}[!t]
\centering
\includegraphics[width=3.5in]{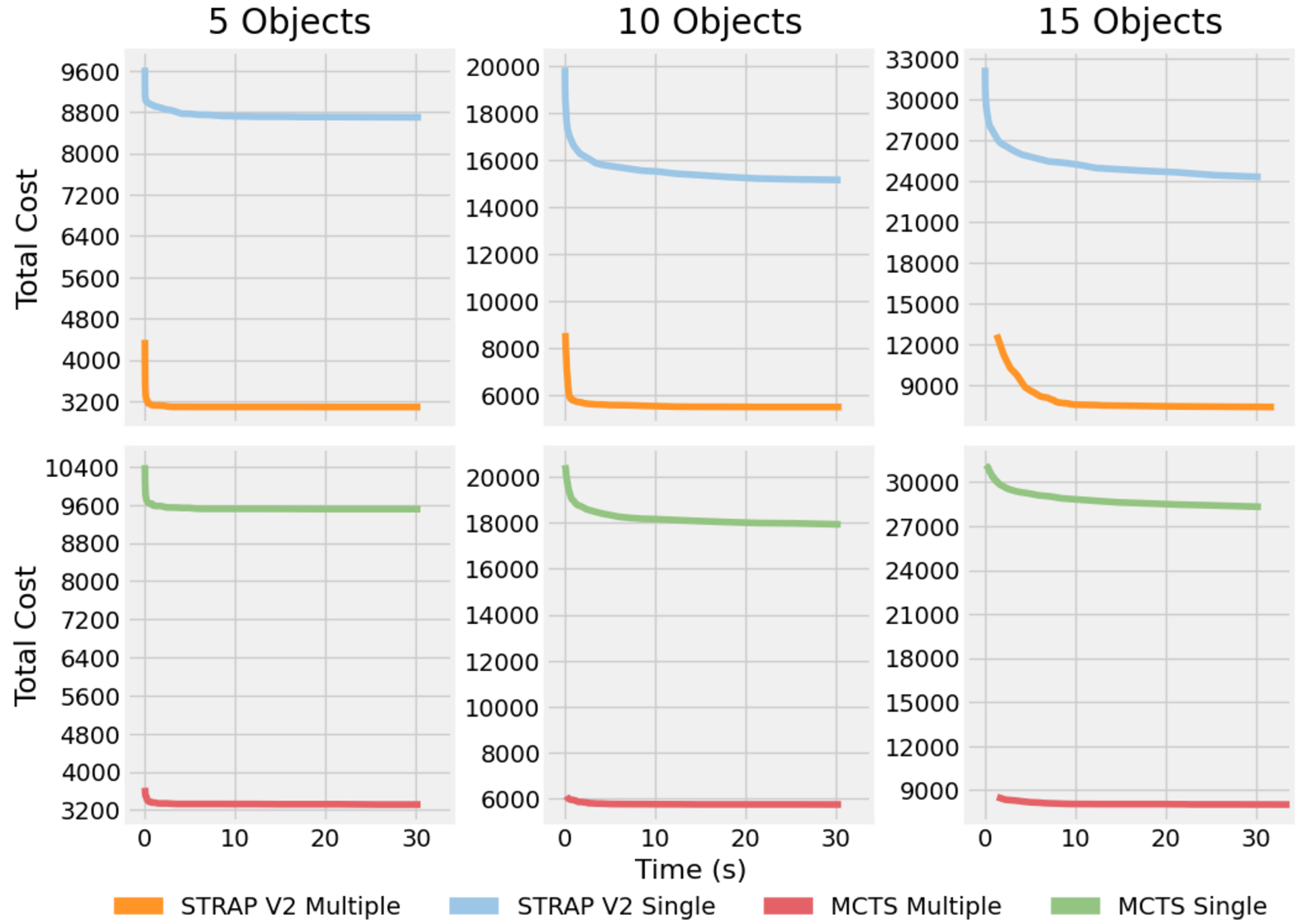}
\caption{\textbf{Comparison of different relocation action strategies} Planners with names ending in ``single'' use the single relocation action strategy, while those ending in ``multiple'' employ the multiple relocation action strategy.}
\label{fig:eval_a}
\end{figure}

This section compares various relocation action strategies across different planners and object counts, as illustrated in Fig.~\ref{fig:eval_a}. Due to the design of ORLA*~\cite{orla}, incorporating multiple-relocation action strategy is impractical. Therefore, we limit our comparison to STRAP V2 and MCTS here.

The results indicate that both the STRAP V2 and MCTS planners, which employ a multiple relocation action strategy, significantly reduce the total cost across all object counts compared to those using a single relocation action strategy, effectively minimizing unnecessary movements. Furthermore, the multiple relocation action strategy enables planners to explore the search space more efficiently with longer operation sequences between states. Consequently, the search tree reaches a goal arrangement state more rapidly, often yielding high-quality solutions in a shorter time.

\subsection{Evaluation on Different Searching Algorithms}\label{eval_b}

This section compares STRAP V2 with previous planners~\cite{trlb, orla, mcts2020labbe, strap2024}. Since ORLA~\cite{orla} cannot incorporate the multiple relocation action strategy, we do not compare planners using multiple relocation action strategy. 

\begin{figure}[H]
\centering
\includegraphics[width=3.5in]{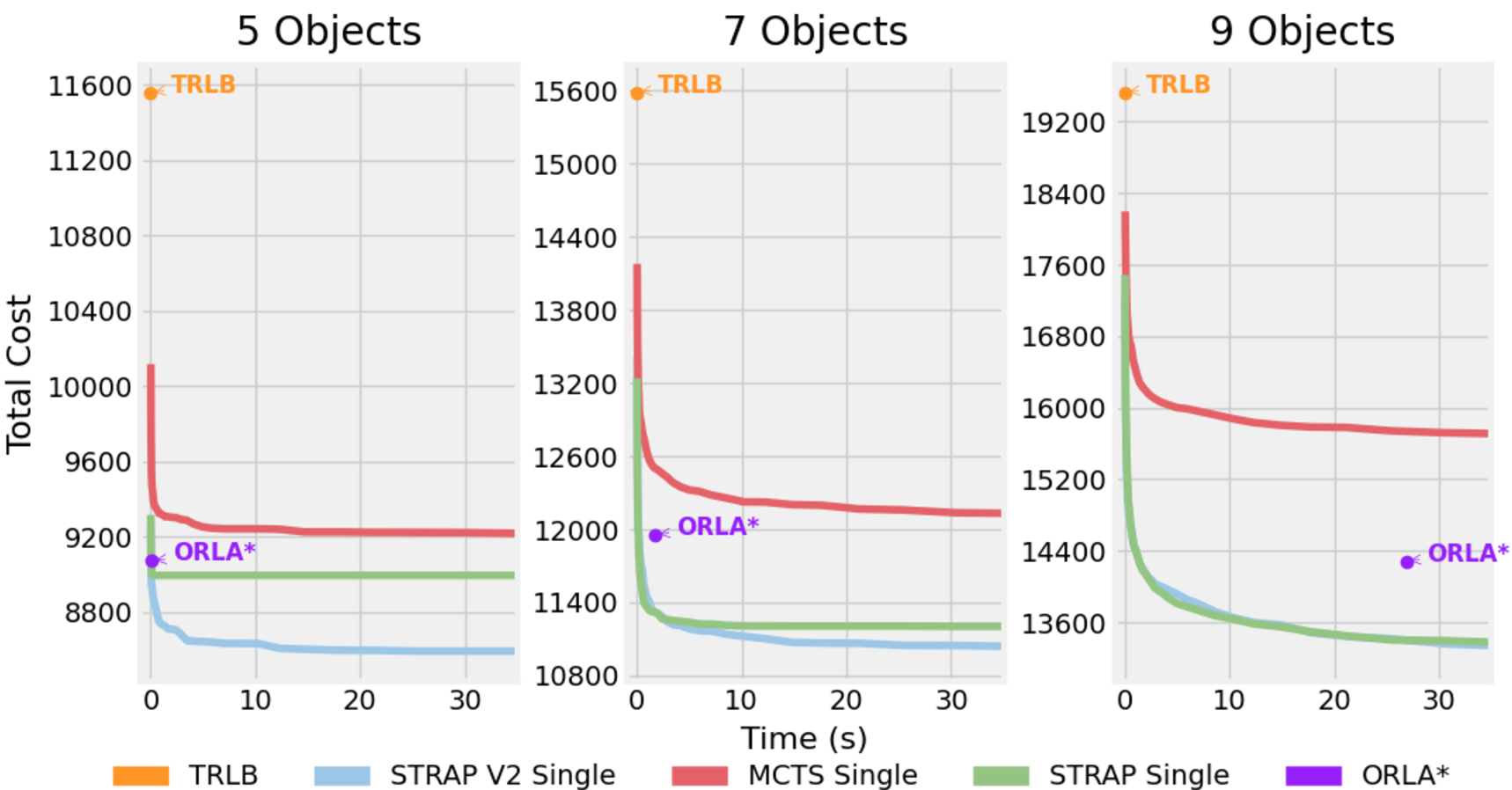}
\caption{\textbf{Comparison of different planners with single relocation action strategy} Both TRLB and ORLA* use single relocation action strategy, so no ``single'' after them.}
\label{fig:eval_b_1}
\end{figure}

Fig.~\ref{fig:eval_b_1} illustrates the performance of all planners using a single-relocation action strategy. Since ORLA* requires extended time to achieve a 100\% success rate, this evaluation includes only cases with 5, 7, and 9 objects. Consistent with prior results in ~\cite{orla, strap2024}, TRLB can produce feasible solutions in short time. Although MCTS yields improved solutions, it fails to converge to a superior solution compared to ORLA*, even with a long planning time. In contrast, STRAP achieves convergence to a higher-quality solution than ORLA* within a shorter planning timeframe. Due to its re-exploration capability, STRAP V2 can continue reducing the total cost over extended planning time, ultimately achieving the highest quality plan among all planners.

\begin{figure}[!h]
\centering
\includegraphics[width=3.5in]{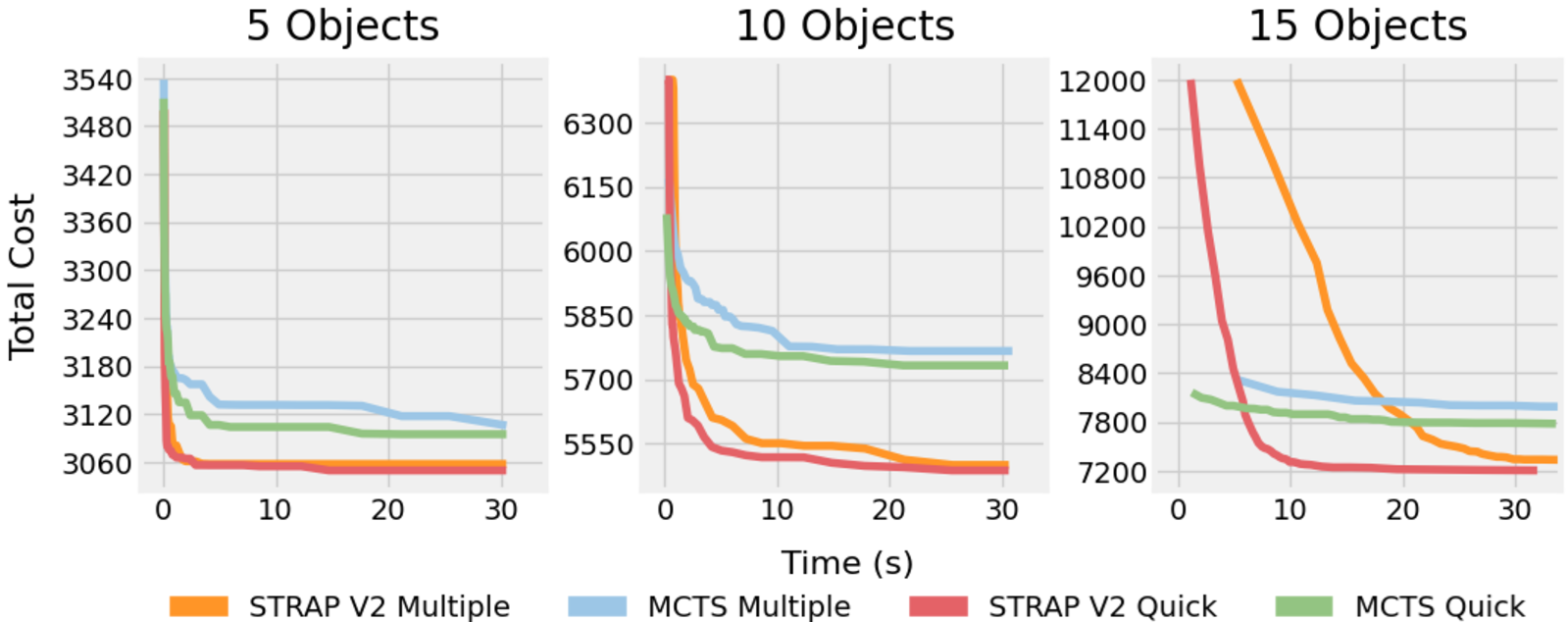}
\caption{\textbf{Comparison of planners using multiple relocation action strategy} Planners with names ending ``Quick'' use multiple relocation action strategy with region reduction strategy.}
\label{fig:eval_b_2}
\end{figure}

\subsection{Evaluation on Region Reduction}\label{state_reduction}
As discussed in Sec.~\ref{sec:multiple_relocation},  an increase in the number of objects results in a corresponding rise in the number of manipulation regions per state, which subsequently increases the number of possible successor states. This, in turn, significantly prolongs the exploration time required for each iteration. For instance, as shown in Fig.~\ref{fig:eval_b_2}, require substantially more time to explore the initial state and find a feasible solution in scenarios with 15 objects. To mitigate this, the planner adopts the region reduction strategy to decrease the number of states requiring exploration. The results indicate that planners incorporating region reduction strategy can find feasible solutions within a shorter time frame and converge to higher-quality solutions more rapidly, even though they ultimately achieve the same solution quality.

\subsection{Ablation Study}
\begin{figure}[!h]
\centering
\includegraphics[width=3.5in]{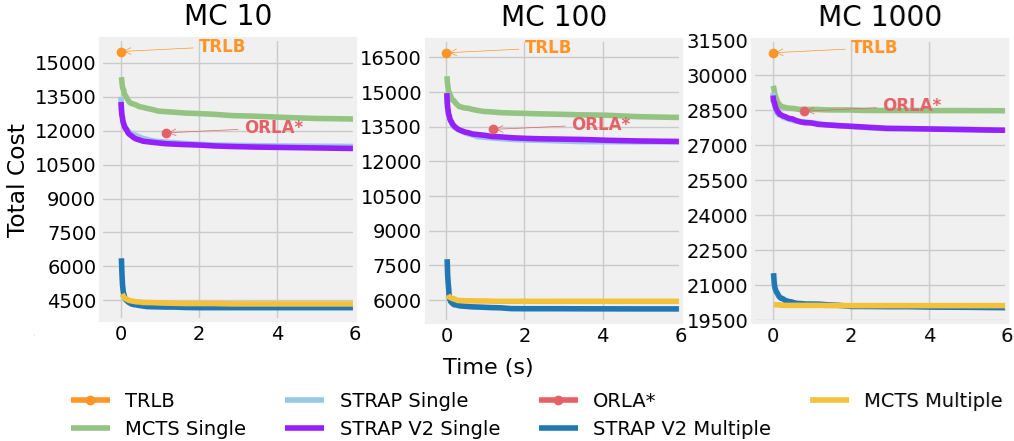}
\caption{\textbf{Comparison planners with higher manipulation costs(MC)}.}
\label{fig:eval_c}
\end{figure}

Fig.~\ref{fig:eval_c}  compares the performance of all planners with different higher manipulation costs. The results demonstrate that both STRAP V2 and the multiple relocation action strategy consistently help the system converge to the highest-quality plan across all manipulation cost scenarios.

\section{Conclusion}
This paper proposes a new A*-based algorithm with a novel rearrangement strategy for mobile robot leveraging multiple relocation actions for improved efficiency. However, our method enhances plan quality over time though optimality cannot be guaranteed. 

\addtolength{\textheight}{-12cm}   









\end{document}